\documentclass[preprint,12pt]{elsarticle}

\usepackage{amssymb}
\usepackage{amsmath}

\usepackage{natbib}
\usepackage{booktabs}
\journal{Expert Systems With Applications}

\begin{document}

\begin{frontmatter}



\title{WRT-SAM: Foundation Model-Driven Segmentation for Generalized Weld Radiographic Testing}

             

\author{Yunyi Zhou, Kun Shi, Gang Hao} 

\affiliation{organization={China Special Equipment Inspection and Research Institute},
            addressline={}, 
            city={Beijing},
            postcode={100029}, 
            state={},
            country={P.R. China}}

\begin{abstract}
Radiographic testing is a fundamental non-destructive evaluation technique for identifying weld defects and assessing quality in industrial applications due to its high-resolution imaging capabilities. Over the past decade, deep learning techniques have significantly advanced weld defect identification in radiographic images. However, conventional approaches, which rely on training small-scale, task-specific models on single-scenario datasets, exhibit poor cross-scenario generalization. Recently, the Segment Anything Model (SAM), a pre-trained visual foundation model trained on large-scale datasets, has demonstrated exceptional zero-shot generalization capabilities. Fine-tuning SAM with limited domain-specific data has yielded promising results in fields such as medical image segmentation and anomaly detection.
To the best of our knowledge, this work is the first to introduce SAM-based segmentation for general weld radiographic testing images. We propose WRT-SAM, a novel weld radiographic defect segmentation model that leverages SAM through an adapter-based integration with a specialized prompt generator architecture. To improve adaptability to grayscale weld radiographic images, we introduce a frequency prompt generator module, which enhances the model’s sensitivity to frequency-domain information. Furthermore, to address the multi-scale nature of weld defects, we incorporate a multi-scale prompt generator module, enabling the model to effectively extract and encode defect information across varying scales.
Extensive experimental evaluations demonstrate that WRT-SAM achieves a recall of 78.87\%, precision of 84.04\%, and an AUC of 0.9746, setting a new state-of-the-art (SOTA) benchmark. Moreover, the model exhibits superior zero-shot generalization performance, highlighting its potential for practical deployment in diverse radiographic testing scenarios.
\end{abstract}



\begin{keyword}


weld radiographic testing, image recognition, defect segmentation, segment anything model, prompt generator
\end{keyword}

\end{frontmatter}



\section{Introduction}
\label{sec1}

Welded structures are extensively utilized in industrial applications, particularly in the construction of pressure vessels (e.g., boilers and storage tanks) and pressure pipelines (e.g., gas and oil transportation systems)~\cite{li2023defect, zuo2023x}. These critical infrastructures often operate under extreme conditions, including high temperatures and pressures, while also containing flammable or hazardous substances. Any structural failure or leakage can lead to catastrophic industrial accidents, resulting in severe economic losses and posing significant threats to human safety and environmental integrity. Consequently, ensuring the structural integrity and reliability of welded components is of paramount importance.

Radiographic testing (RT) is one of the most widely adopted NDT (non-destructive testing) methods for weld defect detection due to its ability to capture high-resolution internal structures and provide intuitive visual representations of defects~\cite{anouncia2006non,galos2021novel}. By analyzing RT images, engineers can identify defect locations, classify defect types, and assess their severity, which is essential for maintaining the quality and safety of welded components. With the increasing scale of industrial production and the corresponding rise in RT inspection workloads, there is a growing demand for enhanced efficiency and accuracy in defect evaluation. However, conventional manual inspection methods suffer from limitations such as low efficiency, subjectivity, poor reproducibility, and difficulties in standardization, leading to inconsistencies in defect assessments.

Recent advancements in computer vision and artificial intelligence have enabled the automation of defect detection in RT images, significantly improving evaluation accuracy and efficiency. Early research~\cite{hu2023bag, yan2024enhancing} primarily relied on traditional image processing techniques and machine learning models, which required handcrafted feature extraction and lacked adaptability across diverse scenarios. With the advent of deep learning, convolutional neural networks (CNNs) have become the dominant approach, achieving remarkable performance in defect classification, segmentation, and detection~\cite{li2023defect,zuo2023x,palma2024deep,wang2024new,ajmi2024advanced,perri2023welding}. However, these CNN-based methods are typically trained on limited datasets, making them prone to overfitting and restricting their generalization to unseen scenarios. 

The recent emergence of large-scale vision foundation models, exemplified by the  Segment Anything Model (SAM) ~\cite{kirillov2023segment}, has introduced a paradigm shift in visual understanding. Pre-trained on massive datasets, these models exhibit strong zero-shot generalization capabilities and have demonstrated outstanding performance in various domains, including medical image segmentation and anomaly detection. Given their scalability and adaptability, fine-tuning vision foundation models for domain-specific applications has become an effective strategy for improving model robustness in complex industrial settings. However, the direct application of such models to RT-based weld defect identification presents unique challenges, primarily due to the grayscale nature of RT images and the multi-scale characteristics of welding defects.

Unlike natural scene images, which contain rich color and texture information, RT images are grayscale, limiting the features that can be captured by standard CNN-based feature extractors. Existing deep learning approaches primarily rely on spatial domain feature extraction, while neglecting valuable frequency-domain information. Inspired by advancements in medical imaging, where frequency-domain analysis has been successfully incorporated to enhance segmentation performance~\cite{wu2024medsegdiff,yu2025improving}, we introduce a  Frequency Prompt Generator (FPG)  to improve SAM's adaptability to grayscale RT images. By leveraging frequency-domain information, our model enhances defect feature representation and improves segmentation accuracy.

Furthermore, welding defects exhibit significant variability in shape, size, and distribution, necessitating a model capable of extracting multi-scale features. Traditional CNN-based methods often struggle with multi-scale defect representation due to fixed receptive field constraints. To address this limitation, we introduce a  Multi-Scale Prompt Generator (MSPG) , which enables our model to effectively capture defect features at varying scales, thereby improving segmentation performance across diverse defect types.

In this paper, we propose WRT-SAM (Weld Radiographic Testing - Segment Anything Model), a novel segmentation framework that integrates SAM with a prompt generator architecture via an adapter mechanism. Our contributions can be summarized as follows:
\begin{itemize} 
    \item To the best of our knowledge, this is the first study to introduce a SAM-based segmentation model for general weld RT image segmentation.  
    \item We propose a prompt generator architecture, consisting of the Frequency Prompt Generator (FPG) and the Multi-Scale Prompt Generator (MSPG), which enhances SAM’s ability to process grayscale RT images and improves its capability to extract defect features across multiple scales.  
    \item We conduct extensive evaluations on three weld RT image segmentation datasets, including GDXray and a proprietary dataset. WRT-SAM achieves state-of-the-art (SOTA) performance, demonstrating strong zero-shot generalization  across different real-world scenarios.  
\end{itemize}

The remainder of this paper is structured as follows: Section~\ref{sec2} reviews related work on RT-based weld defect detection and vision foundation models. Section~\ref{sec3} presents the proposed  WRT-SAM  framework and its key components. Section~\ref{sec4.1} to ~\ref{sec4.3} details the experimental setup, including dataset descriptions, evaluation metrics, and implementation details. Section~\ref{sec4.4} and  ~\ref{sec4.5} discusses the experimental results and comparative analyses. Finally, Section~\ref{sec5} concludes the paper and outlines potential directions for future research.

\section{Related Work}
\label{sec2}

\subsection{Deep Learning-Based Weld Radiographic Image Recognition}  
\label{subsec2.1}
Recent advancements in deep learning have significantly improved the recognition of weld defects in radiographic testing images. Various approaches have been proposed to enhance defect detection, particularly in handling small targets, capturing low-sensitivity spatial information, and optimizing segmentation accuracy across multi-scale defects.
YOLOV5 ~\cite{xu2023defect} enhances small-target detection and spatial feature extraction by integrating the coordinate attention (CA) mechanism, SIOU loss function, and FReLU activation function, facilitating global optimization for defect detection. Similarly, Improved-U-Net~\cite{yang2021automatic} introduces additional skip connections between encoder-decoder layers, mitigating information bottlenecks and improving segmentation performance on multi-scale welding defects. 
To further refine small-scale defect identification, MAU-Net~\cite{wang2024new} incorporates a convolutional block attention mechanism, optimizing large-scale feature extraction through multi-scale even convolution. Meanwhile, the multiple scale spaces (MSS)-empowered segmentation method~\cite{liu2023multiple} addresses the scale variability challenge by constructing three feature spaces: (1) a multi-scale feature space using dilated convolutions, (2) a multi-scale semantic space via max pooling with varying window sizes, and (3) a multi-scale relational space through a self-attention mechanism. 

Despite the significant progress in CNN-based segmentation architectures, existing studies predominantly focus on improving defect detection accuracy within specific scenarios, often neglecting cross-scenario generalization. As a result, these models exhibit limited adaptability to the complex and diverse real-world conditions encountered in industrial radiographic testing.

\subsection{Visual Tuning on SAM for Downstream Tasks}  
\label{subsec2.2}

Visual tuning techniques for adapting pre-trained models can be broadly classified into fine-tuning, prompt tuning, adapter tuning, parameter tuning, and remapping tuning~\cite{yu2024visual}. Among these, prompt tuning and adapter tuning offer an efficient means to transfer pre-trained models to domain-specific applications. 
Segment Anything Model (SAM)~\cite{kirillov2023segment}, a foundational model for image segmentation, leverages prompt-based adaptation for diverse segmentation tasks. Several approaches have been proposed to refine its domain-specific performance. PA-SAM~\cite{xie2024pa} enhances segmentation accuracy by refining mask decoder features at multiple prompt levels. SSPrompt-SAM learns spatial and semantic prompts through adaptive weighting, improving domain-specific adaptation. SAM-Adapter~\cite{chen2023sam}, a lightweight extension using multi-layer perceptrons (MLPs), effectively injects task-specific knowledge into SAM. RobustSAM~\cite{chen2024robustsam} addresses image degradation issues while preserving SAM’s zero-shot learning capabilities through the anti-degradation output token generation (AOTG) and anti-degradation mask feature generation (AMFG) modules.
Given the effectiveness of prompt and adapter tuning, recent studies in pose-guided generation and virtual dressing~\cite{shen2024imagpose, shen2024imagdressing} have demonstrated the potential of foundation models in structured generation tasks. Additionally, works on conditional diffusion models~\cite{shen2023advancing, shen2024boosting} highlight the importance of rich contextual information in enhancing generative performance. 

Motivated by these advancements, we integrate both tuning strategies to optimize WRT-SAM for weld radiographic defect segmentation, ensuring its robustness and adaptability in non-destructive testing applications.

\section{Proposed Method}
\label{sec3}

\subsection{Overview}
\begin{figure*}[t]
    \centering
    \includegraphics[width=1\linewidth]{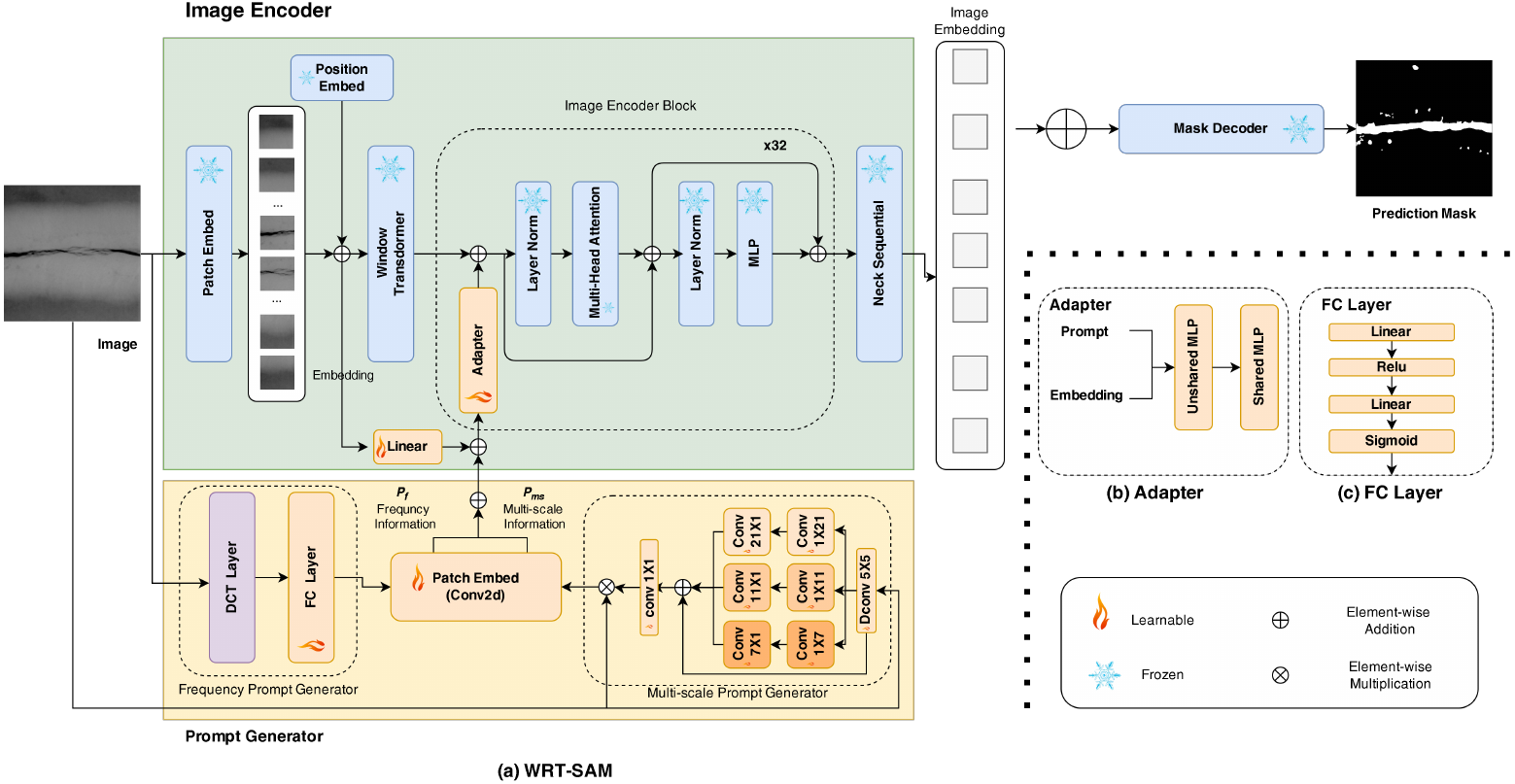}
\caption{\fontsize{11}{11}\selectfont Overview of our proposed WRT-SAM framework.}
\label{fig:WRT-SAM}
\end{figure*}
The framework of WRT-SAM proposed in this paper, as shown in Fig. \ref{fig:WRT-SAM}, consists mainly of three parts: an adaptable Image Encoder, a frequency prompt generator, and a multi-scale prompt generator. Ultimately, these pieces of information are summed and injected into a frozen Mask Decoder together to generate the final segmentation mask.
\subsection{Frequency Prompt Generator}  
The FcaNet\cite{qin2021fcanet} network has demonstrated exceptional performance in enhancing existing models for object detection and instance segmentation tasks by extracting features from frequency domain channel information using Discrete Cosine Transformation (DCT). This is particularly beneficial for grayscale images, such as X-ray inspection images, where frequency domain information helps differentiate between low-frequency content (typically the main subject) and high-frequency content (often noise or fine details), thereby increasing the prominence of defect-related information. Consequently, we have integrated an FcaNet-like structure into the frequency prompt generator component of our WRT-SAM network architecture.

Specifically, the original input image $X$
 is divided into patches, which are then fed into the DCT layer. In the DCT layer, corresponding filters are first derived based on a predefined quantization table. A 2D DCT transformation is then applied to the patched images. The transformed images are then passed through a fully connected layer (FC layer) for dimensional mapping to extract frequency information. Finally, the frequency information $P_f$ is encoded to generate the frequency prompts.
\begin{equation}
P_f=Conv2d(fc(MSCDCT(X))).
\label{eq:frequency prompt generator} 
\end{equation}
where $X \in \mathbb{R}^{C\times H\times W} $  is the input image, $MSCDCT$ is the multi spectral channel Discrete Cosine Transformation which can be further expressed in Eqs. \ref{eq:MSCDCT}, $fc$ denotes the mapping functions like fully connected layer, Conv2d is the embedding layer to generate the frequency prompts.
\begin{equation}
MSCDCT(X)=cat([Freq^0,Freq^1,...,Freq^{n-1}],
\label{eq:MSCDCT} 
\end{equation}
where the input $X$ is split into many parts along the channel dimension. Denote $[X^0,X^1,...,X^{n-}]$, in which $X^i \in \mathbb{R}^{C'\times H\times W} $, $i \in {0,1,...,n-1}, C'=\frac{C}{n} $, and $C$ should be divisible by $n$. For each part, a corresponding 2D DCT results can be computed as,
\begin{equation}
\begin{split}
 Freq^i=2DDCT^{u_i,v_i}(X^i),\\
=\sum_{h=0}^{H-1}\sum_{\omega=0}^{W-1}X_{:h,\omega}^{i}B_{h,\omega}^{u_i,v_i} 
\end{split}
\label{eq:DCT} 
\end{equation}

\begin{center}
$s.t. i \in {0,1,...,n-1},$
\end{center}

\begin{equation}
B_{h,\omega}^{u_i,v_i}=cos(\frac{\pi h}{H}(u_i+\frac{1}{2}))cos(\frac{\pi \omega}{W}(v_i+\frac{1}{2})).
\label{eq:basis DCT} 
\end{equation}
in which $[u_i,v_i]$ are the frequency component 2D indices corresponding to $X^i$, and $Freq^i \in \mathbb{R}^{C'}$, $B_{h,\omega}^{u_i,v_i}$ is the basis function of 2D DCT.

\subsection{Multi-scale Prompt Generator}  
In the multi-scale prompt generator component of our WRT-SAM network architecture, SegNeXt\cite{guo2022segnext} demonstrates that convolutional attention is a more efficient and effective method for encoding contextual information than the self-attention mechanism in transformers. This is particularly relevant considering that the defect scales in weld radiographic testing images span a multidimensional range of small, medium, and large targets, as defined by MS COCO\cite{MSCOCO}. Consequently, the network must possess robust multi-scale feature extraction capabilities. Thus, we introduce the Multi-scale Convolutional Attention (MSCA) mechanism as the multi-scale prompt information extraction module, implemented as the multi-scale prompt generator shown in the diagram.

Specifically, the original input image first passes through a Dconv layer to fuse local information, where Dconv refers to depthwise convolution. Subsequently, multi-branch depthwise strip convolutions are employed to capture multi-scale contextual information. Finally, a $1\times 1$ convolution is applied to integrate information across different scale channels, resulting in the final multi-scale prompt $P_{ms}$. The process described above can be defined as follows:
\begin{equation}
P_{ms}=(Conv_{1X1}(\sum_{i=0}^{3}Scale_i(Dconv(X))))\bigotimes X.
\label{eq:multi-scale prompt generator} 
\end{equation}
where $X \in \mathbb{R}^{C\times H\times W} $  is the input image, $\bigotimes$ is the elements-wise multiplication, $Dconv$ denotes depth-wise convolution and $Scale_i, i\in \left\{0,1,2,3 \right\}$, denotes the $i$th branch. $Scale_0$ is the identity connection. In each branch, there are two depth-wise strip convolutions to approximate standard depth-wise convolutions with large kernels. Here, the kernel size for each branch is set to 7, 11, and 21, respectively. $P_{ms}$ denotes the multi-scale prompt.

\subsection{Loss Function}  
When training the WRT-SAM, we choose to use the IOU (Intersection over Union) loss function to calculate the loss incurred by the deviation between the predicted value and the true value. The formula is as follows:
\begin{equation}
L_{IoU}\text{ = }1 \text{ — } IoU.
\label{eq:IOU-loss} 
\end{equation}

\begin{equation}
IoU\text{ = }\frac{Intersec}{Union}.
\label{eq:IOU} 
\end{equation}
where $Intersec$ and $Union$ are the intersection and union between precision and the Ground Truth respectively.

\section{Experiment and Analysis}
\label{sec4}
To validate the superiority and generalization of the proposed WRT-SAM method, it is compared with multiple state-of-the-art segmentation approaches on two datasets, namely, GDXray and our private dataset.

\subsection{Datasets}\label{sec4.1}
\textbf{\emph{GDxray}}~\cite{carrasco2023gdxray} is a large dataset consisting of five image groups, each designed for different applications (i.e., \textit{Castings, Welds, Baggage, Nature, and Settings}). The \textit{Welds} category contains 68 weld radiographic test images, of which only 10 are provided with official segmentation annotations. We refer to these as GDXray-10, while the remaining 58 images are named GDXray-58.

\textbf{\emph{Our private dataset, WRTD}} contains 115 weld radiographic images collected from different projects and equipment. The welds were made using various materials and welding processes, and the radiographic images were obtained through different inspection methods. These factors contribute to variations in data distribution and increase the observable difficulty of detecting weld and defect features in the images.

\subsection{Evaluation Metrics} \label{sec4.2}
To better verify the performance of proposed network model, some evaluation indicators are introduced here to verify the location results of welding defects.

Precision($P$) and Recall($R$) are two important indicators of artificial intelligence to evaluate the neural network models, as shown in Eqs. \ref{eq:precision} and \ref{eq:recall}. 
\begin{equation}
P=\frac{T_p}{T_p+F_p}.
\label{eq:precision} 
\end{equation}
\begin{equation}
R=\frac{T_p}{T_p+F_n}.
\label{eq:recall} 
\end{equation}
where $T_p$ and $T_n$ denote true positive and true negative. $F_p$ and $F_n$ denote false positive and false negative.

Area under Precision-Recall Curve ($AUC$): The value of $AUC$ represents the area under the curve, such as $P$-$R$ Curve, which is between 0 and 1. It could evaluate the performance of network intuitively. The bigger the $AUC$ value, the better the performance of the model.

IoU can also measure the similarity between the predicted region and the real region, as in Eqs. \ref{eq:IOU}.

\subsection{Implementation Details}\label{sec4.3} 

For training our WRT-SAM, We use the AdamW optimizer, the learning rate is set to $0.0002$, the minimum learning rate is $\exp{^{-7}}$, the epoch is $20$. For the input images, we remain the hight of the original images and crop them through their width to make the input images have the same width of 640 pixels on the GDXray dataset.

\subsection{Comparison with State-of-the-art Methods} \label{sec4.4}

\subsubsection{Comparisons on GDXray}
As shown in Table \ref{tbl:SOTA-compare-GDXray}, we trained our model on the GDXray dataset and compared it with several classic state-of-the-art (SOTA) segmentation algorithms and a baseline algorithm. Our WRT-SAM model achieved recall and precision rates of 78.87\% and 84.04\%, respectively, outperforming the other algorithms in comparison. Additionally, its Area Under the Curve (AUC) was 1.6\% higher than that of the baseline. These results demonstrate that the SAM-based defect segmentation model has the potential to surpass existing SOTA algorithms and achieve improved performance in weld radiographic inspection image analysis.
\begin{table}[t]
	\centering
	\caption{\fontsize{11}{11}\selectfont the weld defect segmentation results on gdxray dataset in comparison with state-of-the art methods}
	\begin{tabular}{llllll} 
		\toprule
		\textbf{Methods} & \textbf{Recall} & \textbf{Precision} & \textbf{AUC} & \textbf{IoU}\\
        \hline
		\midrule
 
		U-Net &  74.34 & 75.97  & $\times$ & 66.72   \\
		Deeplabv3 & 78.36 & 79.52  & $\times$ &  \textbf{73.54}  \\
		PspNet & 77.11 & 78.75  & $\times$ & 71.23   \\
		U-Net+CBMA & 77.33 & 78.43  & $\times$ & 71.12 \\
            SAM-Adapter(Baseline) &78.87 &78.39 &0.9596 &49.25\\
            \textbf{Ours}&\textbf{78.87} & \textbf{84.04} & \textbf{0.9746}  & 51.36 \\
            
		\bottomrule
	\end{tabular}
	\label{tbl:SOTA-compare-GDXray}
\end{table}

\subsubsection{Comparisons on private dataset}
To demonstrate that our proposed method is better suited for practical application scenarios, we trained the model on our own dataset of 115 images and compared the results with those of the U-Net++ network and the baseline method. As shown in Table \ref{tbl:SOTA-compare-private-dataset}, our method achieved a recall rate of 79.61\%, which is 11.13\% and 2.18\% higher than U-Net++ and the baseline, respectively. This is significant for weld quality management in real-world production scenarios. Additionally, although the SAM-based methods underperformed compared to U-Net++ in terms of the IoU metric, our WRT-SAM showed a 2.38\% improvement over the baseline method, demonstrating the effectiveness of our approach.
\begin{table}[t]
	\centering
	\caption{\fontsize{11}{11}\selectfont the weld defect segmentation results on private dataset in comparison with state-of-the art methods}
	\begin{tabular}{llllll} 
		\toprule
		\textbf{Methods} & \textbf{Recall} & \textbf{Precision} & \textbf{AUC} & \textbf{IoU}\\
        \hline
		\midrule
 
		U-Net++ &  71.64 & \textbf{83.42} & $\times$ & \textbf{63.05}   \\
        SAM-Adapter(Baseline) &77.91 &78.98  &0.9724 &57.96\\
        \textbf{Ours}&\textbf{79.61} & 77.70 & \textbf{0.9796}  & 59.34 \\
            
		\bottomrule
	\end{tabular}
	\label{tbl:SOTA-compare-private-dataset}
\end{table}

\subsection{Ablation Studies and Analysis}\label{sec4.5}  

The comparison results presented in Table \ref{tbl:SAM-Adapter}, Table  \ref{tbl:FcaNet}, Table \ref{tbl:MSCA}, \ref{tbl:GDXray-58} and Table \ref{tbl:Private_clean}, demonstrate that the proposed WRT-SAM method is superior to the foundation SAM and baseline method SAM Adapter~\cite{chen2023sam}. In what follows,  the proposed WRT-SAM method is comprehensively analyzed from 4 aspects to investigate the logic behind its superiority.

\subsubsection{ Role of the SAM adapter}

The purpose of this experiment is to assess the adaptability of the SAM (Segmentation Anything Model) for weld radiographic testing image segmentation. We use 10 images with official defect annotations from the welds category in the GDXray dataset (GDXray-10) for training, validation, and testing. Using the 'Segment Everything' mode of the foundation SAM model, we obtain segmentation masks and compare them with the official ground truth to calculate accuracy metrics. For the SAM adapter, we split the dataset into a training set and a validation set in an 8:2 ratio, freeze the base model, and train only the adapter component. Finally, we evaluate the model's performance on the validation set. The experimental results are presented in Table \ref{tbl:SAM-Adapter}. Compared to the baseline SAM, the SAM adapter demonstrates significant improvements across all metrics, thereby confirming the feasibility of applying SAM to the downstream task of weld radiographic testing image defect segmentation through adapter-based tuning.
\begin{table}[t]
    \centering
	\caption{\fontsize{11}{11}\selectfont ablation studies on role of adapter}
	\begin{tabular}{llllll} 
		\toprule
		\textbf{Methods} & \textbf{Recall} & \textbf{Precision} & \textbf{AUC} & \textbf{IoU}\\
        \hline
		\midrule

            SAM(Everything mode) & 39.79 & 49.18  & $\times$  & 1.13 \\
            SAM Adapter\text{(Baseline)} &\textbf{78.87} &\textbf{78.39} &\textbf{0.9596} &\textbf{49.25}\\
		\bottomrule
	\end{tabular}
	\label{tbl:SAM-Adapter}
\end{table}

\subsubsection{Influence of frequency prompt generator}
The objective of this experiment is to introduce the Frequency Prompt Generator (FPG) module and evaluate its impact on defect segmentation performance. We continue to use the previously mentioned GDXray-10 dataset for training and validation. Building on the baseline model, we integrate the FPG module and, by adjusting the frequency ranges corresponding to the basis functions of the Discrete Cosine Transform (DCT) filter, select DCT parameters that yield optimal performance. The specific experimental results are presented in Table \ref{tbl:FcaNet}. After incorporating the FPG module, the model's precision, AUC, and IoU all improve, while the recall decreases slightly. This is primarily due to the fact that DCT is a compression process, during which some image details are lost. After comprehensively comparing all metrics, we select the 'top 1' mode parameters (where 
$[u_i,v_i]$ are both set to 0) for application in the final WRT-SAM model.

\begin{table}[t]
	\centering
	\caption{\fontsize{11}{11}\selectfont ablation studies on influence of frequency prompt generator}
	\begin{tabular}{llllll} 
		\toprule
		\textbf{Methods} & \textbf{Recall} & \textbf{Precision} & \textbf{AUC} & \textbf{IoU}\\
        \hline
		\midrule

            SAM Adapter(Baseline) &\textbf{78.87} &78.39 &0.9596 &49.25\\
            \textbf{SAM Adapter+FPG(top1)} &78.32 & 81.87 &\textbf{0.9859} &\textbf{51.87} \\
            \textbf{SAM Adapter+FPG(bot1)} &77.10 & \textbf{83.66} & 0.9779 & 51.03 \\
		\bottomrule
	\end{tabular}
	\label{tbl:FcaNet}
\end{table}

\subsubsection{ Impact of multi-scale prompt generator}
The experimental setup for this section is similar to the ablation studies on the influence of the frequency prompt generator, with the aim of evaluating the impact of the multi-scale prompt generator (MSPG) module on defect segmentation accuracy. The specific results are presented in Table \ref{tbl:MSCA}. The experimental results clearly show that after adding multi-scale information as prompts and integrating them into the baseline network, the model's performance is comprehensively improved, highlighting the positive impact of the MSPG module. Therefore, we retain the MSPG module in the final WRT-SAM model.
\begin{table}[t]
	\centering
	\caption{\fontsize{11}{11}\selectfont ablation studies on impact of MSPG}
	\begin{tabular}{llllll} 
		\toprule
		\textbf{Methods} & \textbf{Recall} & \textbf{Precision} & \textbf{AUC} & \textbf{IoU}\\
        \hline
		\midrule

            SAM Adapter(Baseline) &78.87 &78.39 &0.9596 &49.25\\
            \textbf{SAM-Adapter\text{+MSPG}}&\textbf{79.29}& \textbf{82.14} & \textbf{0.9837} & \textbf{50.88} \\

		\bottomrule
	\end{tabular}
	\label{tbl:MSCA}
\end{table}

\subsubsection{(Zero-shot) generalization analysis}
The experimental results in this section are crucial for practical applications, as they evaluate the model's generalization performance on unknown datasets in future scenarios. We use GDXray-10 as the training set and test the model on GDXray-58 and a private dataset to simulate unknown scenarios the model may encounter in future applications. The experimental results are presented in Table \ref{tbl:GDXray-58} and Table \ref{tbl:Private_clean}, respectively. In Table \ref{tbl:GDXray-58}, our WRT-SAM model achieves the best AUC performance among the four models and outperforms the baseline in both recall and precision, particularly in recall, which is crucial for identifying defects related to the safe operation of equipment. Table \ref{tbl:Private_clean} demonstrates the true zero-shot generalization of the model, illustrating its performance across data from different scenarios, equipment, materials, welding processes, and radiography techniques. The results indicate that our model significantly outperforms the baseline in recall and AUC, particularly in recall, which is critical for practical applications. The IoU has shown slight improvement, while precision remains lower than the baseline. Additionally, the zero-shot generalization performance of the MSPG module is generally strong, which we attribute to the diversity of defect sizes across scenarios. In summary, the WRT-SAM model demonstrates enhanced generalization, critical for its widespread future application.
\begin{table}[t]
	\centering
	\caption{ \fontsize{11}{11}\selectfont generalization analysis on gdxray-58}
	\begin{tabular}{llllll} 
		\toprule
		\textbf{Methods} & \textbf{Recall} & \textbf{Precision} & \textbf{AUC} & \textbf{IoU}\\
        \hline
		\midrule

            SAM Adapter\text{(Baseline)} &9.31 &42.66 &0.5810 &\textbf{23.51}\\
            SAM Adapter+FPG &41.68& \textbf{50.24} & 0.8970 & 10.92\\
            SAM Adapter+MSPG &\textbf{46.45}& 47.92 & 0.9216 & 16.17 \\
            \textbf{Ours}&45.57& 46.87 &\textbf{0.9288 } & 15.18 \\

		\bottomrule
	\end{tabular}
	\label{tbl:GDXray-58}
\end{table}

\begin{table}[t]
	\centering
	\caption{\fontsize{11}{11}\selectfont zero-shot generalization analysis on private dataset}
	\begin{tabular}{llllll} 
		\toprule
		\textbf{Methods} & \textbf{Recall} & \textbf{Precision} & \textbf{AUC} & \textbf{IoU}\\
        \hline
		\midrule

            SAM Adapter\text{(Baseline)} &14.93 &\textbf{74.03} &0.5610 &8.69\\
            SAM Adapter+FPG &44.32& 58.49& 0.8900 & 8.98\\
            SAM Adapter+MSPG &\textbf{52.15}& 60.14& \textbf{0.9193} & \textbf{9.03} \\
            \textbf{Ours}&46.47& 52.84 &0.9088  & 8.90\\

		\bottomrule
	\end{tabular}
	\label{tbl:Private_clean}
\end{table}

\subsection{Visualization}

\begin{figure*}[t]
    \centering
    \includegraphics[height=10cm,width=7cm]{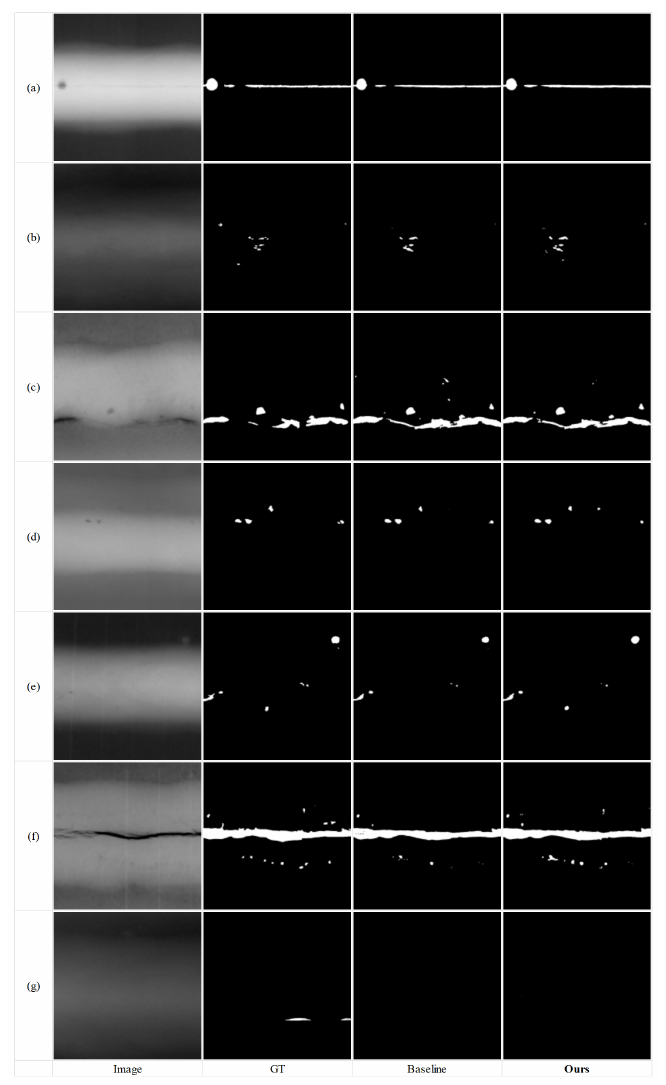}
\caption{\fontsize{11}{11}\selectfont Comparison of baseline SAM-Adapter and our proposed WRT-SAM results visualization on GDXray\text{-10}. From left to right, each column is the original image, the ground truth, the Baseline network predicted results, and our WRT-SAM network predicted results.}
\label{fig:GDXray-10-visual}
\end{figure*}

\begin{figure*}[t]
    \centering
    \includegraphics[height=10cm,width=7cm]{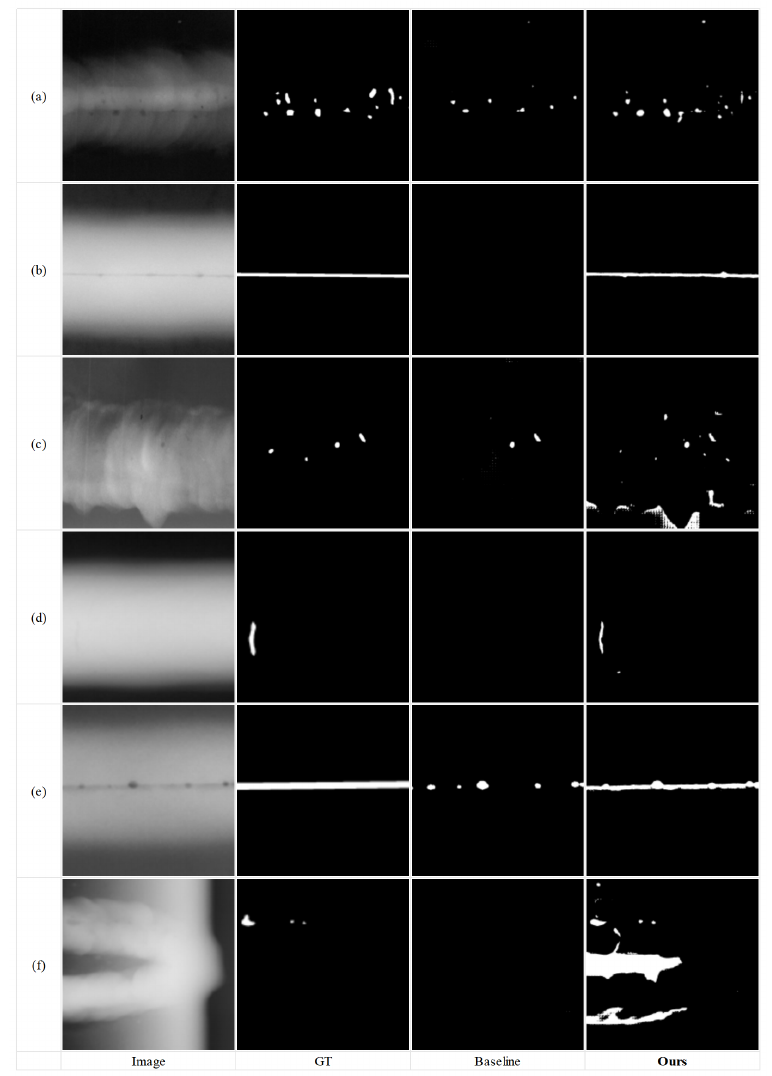}
\caption{\fontsize{11}{11}\selectfont Comparison of baseline SAM-Adapter and our proposed WRT-SAM results visualization on GDXray\text{-58}. From left to right, each column is the original image, the ground truth, the Baseline network predicted results, and our WRT-SAM network predicted results.}
\label{fig:GDXray-58-visual}
\end{figure*}

\begin{figure*}[t]
    \centering
    \includegraphics[height=10cm,width=7cm]{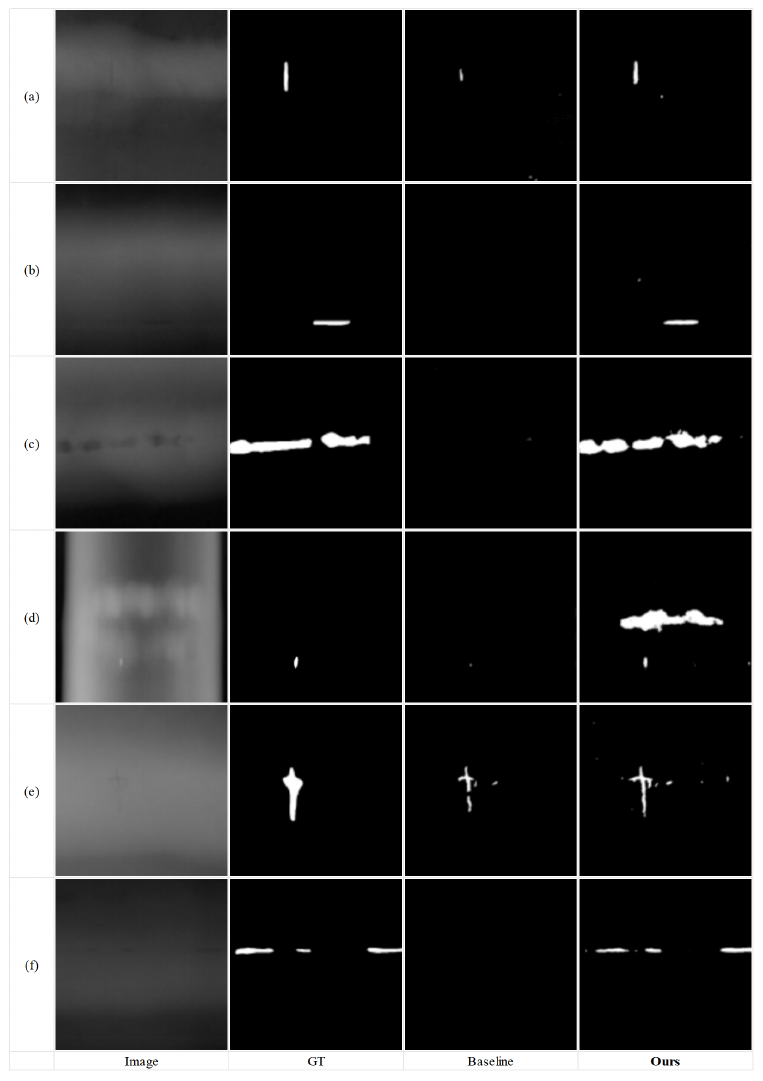}
\caption{\fontsize{11}{11}\selectfont Comparison of baseline SAM-Adapter and our proposed WRT-SAM results visualization on our private dataset. From left to right, each column is the original image, the ground truth, the Baseline network predicted results, and our WRT-SAM network  predicted results.}
\label{fig:private-dataset-visual}
\end{figure*}

 We present and compare the test results of ground truth, Baseline, and our WRT-SAM in Figures \ref{fig:GDXray-10-visual}, \ref{fig:GDXray-58-visual} and \ref{fig:private-dataset-visual}, respectively.
 
The visualization results presented in Fig. \ref{fig:GDXray-10-visual} correspond to the validation outcomes of the WRT-SAM and baseline algorithms in the 'Comparisons on GDXray' section of the 'Comparison with State-of-the-art Methods' part, illustrating the performance of WRT-SAM under a relatively consistent data distribution. Overall, in rows (a) to (f) of Fig. \ref{fig:GDXray-10-visual}, both the baseline and WRT-SAM have successfully segmented the main defects. Furthermore, our method has a clear advantage in handling finer details, capable of identifying defects with lower contrast, as shown in row (f). However, for the defects in row (g), due to the extremely low contrast between the defects and the weld seams, both the baseline and WRT-SAM miss some detections. This also highlights areas for potential improvement in future research.

Fig. \ref{fig:GDXray-58-visual} and \ref{fig:private-dataset-visual} visualize the results of WRT-SAM trained on GDXray-10 and evaluated on GDXray-58 and our own dataset, respectively. These results are presented alongside the ground truth and Baseline algorithm outcomes, corresponding to the "(Zero-shot) generalization analysis" section in the "Ablation Studies and Analysis" chapter.

In rows (b) to (f) of Fig. \ref{fig:GDXray-58-visual}, the baseline algorithm exhibits varying degrees of missed detections, while our WRT-SAM successfully segments the main defects. This suggests that the baseline model fails to handle defect samples from the GDXray dataset that it has not previously encountered. Although our WRT-SAM generates more defect segmentation masks, it still successfully performs defect segmentation on images with somewhat different data distributions, demonstrating the model's strong generalization ability. We were pleased to find that in group (e), our WRT-SAM provided more detailed segmentation results than manual annotation. Future research could focus on methods to filter out redundant segmentation annotations.

The conclusions presented in Fig. \ref{fig:private-dataset-visual} are similar to those in Fig. \ref{fig:GDXray-58-visual}, except that the test data come from a completely different dataset. For entirely unfamiliar data, the baseline model has more missed detections, while our method provides effective segmentation results that capture the correct answers, further demonstrating the zero-shot generalization of our model.

\section{Conclusion}
\label{sec5}

In conclusion, radiographic testing is essential for identifying weld defects and evaluating quality in industrial applications, owing to its non-destructive nature and intuitive imaging characteristics. In the past decade, significant advancements have been made in weld defect identification from radiographic images using machine learning. However, current methods, which involve training small-scale, specialized models on single-scenario datasets, are limited by poor cross-scenario generalization. The pre-trained SAM foundation model, trained on large-scale datasets, exhibits remarkable zero-shot generalization capabilities and has shown promising results in tasks such as medical image segmentation and anomaly detection after fine-tuning with small-scale downstream data. This study introduces the first SAM-based model for general weld radiographic testing image defect segmentation. Based on this study and related research, our proposed method and model have the potential to overcome the limitation of insufficient generalization in practical applications of existing weld radiographic testing image defect segmentation models. To improve the model's adaptability to grayscale weld radiographic images, we introduced a frequency prompt generator module, enhancing the model's focus on frequency domain information. Additionally, we added a multi-scale prompt generator module to address the multi-scale nature of welding defects, improving the model's ability to extract and encode defect information across different scales. Experimental results demonstrated that the WRT-SAM model achieved state-of-the-art performance with a recall of 78.87\%, precision of 84.04\%, and an AUC of 0.9746, while also exhibiting impressive zero-shot generalization capabilities. In the future, our work will focus on: (1) improving the model's detection accuracy for low-contrast defects, (2) enhancing defect segmentation precision while maintaining high recall, and (3) reducing unexpected segmentation.












\bibliographystyle{elsarticle-num} 
\bibliography{ref}
\end{document}